\documentclass[runningheads]{llncs}
\usepackage{graphicx}
\usepackage{booktabs}
\usepackage{float}
\usepackage{subfigure}

\usepackage{CJK}
\usepackage{type1cm}
\usepackage{times}
\usepackage[marginal]{footmisc}

\begin{document}
\title{Exploring Text-transformers in AAAI 2021 Shared Task: COVID-19 Fake News Detection in English}
\titlerunning{COVID-19 Fake News Detection in English using Text-transformers}
%
\author{Xiangyang Li\inst{1*}\orcidID{0000-0003-2862-0239} \and
Yu Xia\inst{1*}\orcidID{0000-0002-8760-4397} \and
Xiang Long\inst{2}\orcidID{2222-3333-4444-5555}
\and Zheng Li\inst{1}\orcidID{0000-0001-5909-3545}
\and Sujian Li\inst{1}
}

\authorrunning{X. Li et al.}

%
\institute{Key Laboratory of Computational 
2
 Linguistics(MOE), Department of Computer Science, Peking University, China\\
\email{\{xiangyangli,yuxia, 1800017744, lisujian\}@pku.edu.cn}
\and
Beijing University of Posts and Telecommunications, China
\email{ xianglong@bupt.edu.cn}\\
}
\maketitle              
\begin{abstract}
In this paper, we describe our system for the  AAAI 2021 shared task of COVID-19 Fake News Detection in English, where
we achieved the 3rd position with the weighted  $F_1$ score of 0.9859 on the test set. Specifically, we proposed an ensemble method of different pre-trained language models such as BERT, Roberta, Ernie, etc. with various training strategies
including warm-up, learning rate schedule and $k$-fold cross-validation. We also conduct an extensive analysis of the samples that are not correctly classified. The code is available at: https://github.com/archersama/3rd-solution-COVID19-Fake-News-Detection-in-

English.

\end{abstract}
\section{Introduction}
Due to the COVID-19 pandemic, offline communication has become less
and tens of millions of people have expressed their opinions and published some news on the Internet. However, some users might publish  some unverified news. If these pieces of news are fake, they may lead to irreparable losses, such as "drinking bleach to kill the new crown virus". Manual detection of these fake news is not feasible because of  huge online communication traffic. In addition, individuals responsible for checking such content may suffer from depression and burnout. For these reasons, it is desirable to build a system that can automatically detect online fake news about COVID-19.

The  Constraint@AAAI 2021 shared task of COVID-19 Fake News Detection in English was organized by 'the First Workshop on Combating Online Hostile Posts in Regional Languages during Emergency Situation'. The data sources are various social media platforms, such as Twitter, Facebook, Instagram, etc. When a piece of social media news is given, the purpose of the shared task is to classify it as fake news or real news.

\footnote{*Equal contribution.}

The rest of the paper is organized as follows:  Section 2 introduces the dataset of this task. Section 3 details
 the architecture of our system (features, models and ensembles). Section 4 offers an analysis of the performance of our models. 
 Section 5 describes the related Work.
 Finally, Section 6 presents our conclusions for this task.

\section{Dataset}
In this section, we first introduce which datasets we use, and perform some exploratory analyses on the dataset.

\subsection{Data Source}
We use the officially provided dataset \cite{patwa2020fighting} and external dataset we collect from the Internet as our training data. The distribution of the data is shown in Table \ref{table1}. 

\begin{table}[]
\caption{Statistics of Datasets.}\label{table1}
\centering
\begin{tabular}{@{}c|c|l|l@{}}
\toprule
Dataset  & Train & Val  & Test \\ \midrule
Official & 6420  & 2140 & 2140 \\ \midrule
External & 699   & 233  & 233  \\ \bottomrule
\end{tabular}
\end{table}

\subsection{Exploratory Data Analysis}

\begin{figure}[ht]
\centering
\caption{The distribution of positive and negative samples in the training and validation set.}
\subfigure[Data distribution in train]{               
\includegraphics[width=4cm,height =5cm]{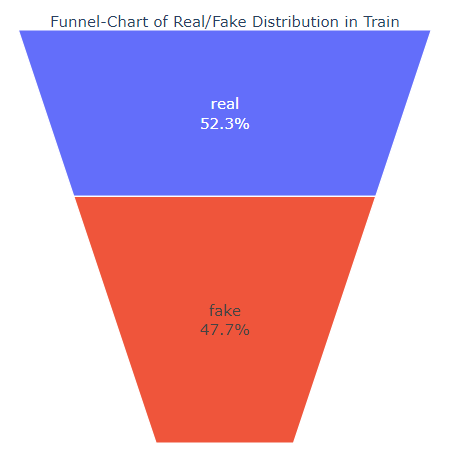}}
\hspace{0in}
\subfigure[Data distribution in valid]{
\includegraphics[width=4cm,height=4.9cm]{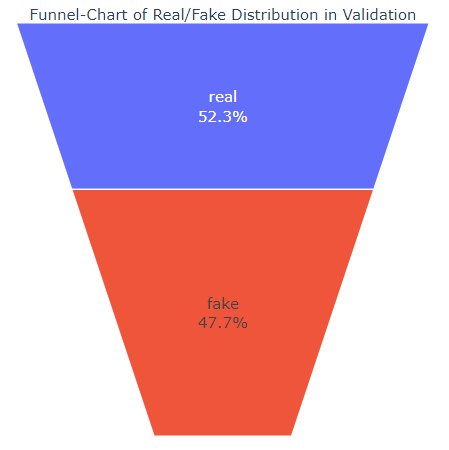}}
\label{figure1}
\end{figure}

In order to have a better understanding of the dataset, we first perform some exploratory analyses on the dataset, which helps us see the hidden laws in the data at a glance and find a model most suitable for the data.

We first explore the distribution of positive and negative samples in the training set and validation set, as shown in Fig.~\ref{figure1}.
From Fig.~\ref{figure1}, we can see that in the training and validation sets, the number of real news exceeds the number of fake news, which illustrates that our dataset is unbalanced, so we can consider a data balanced sampling method when preprocessing data.

\begin{figure}[h]
\centering
\caption{The word cloud diagram of the training set and the validation set. We determine the size of the word in the word cloud according to the frequency of the word.}
\subfigure[Train word cloud]{               
\includegraphics[width=4cm]{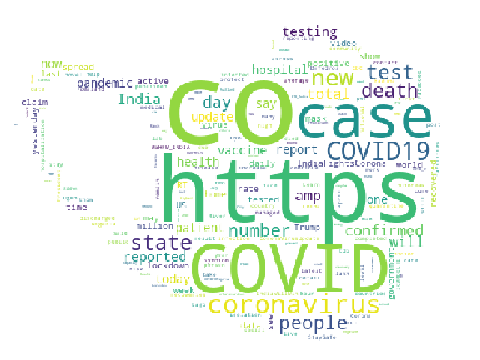}}
\hspace{0in}
\subfigure[Validation word cloud]{
\includegraphics[width=4cm]{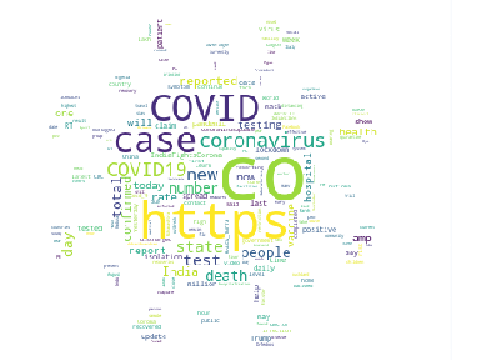}}
\label{figure2}
\end{figure}

In order to analyze the characteristics of the words in the sentence, we  calculate the word frequencies of the training and validation set respectively, remove the stop words, and make the corresponding word cloud diagram as shown in Fig.~\ref{figure2}.

From the Fig.~\ref{figure2}, we can see that 'COVID', 'https', and 'co' are the words with the highest frequency in the dataset. 'COVID', and 'co' appear more frequently than in other normal text, while the higher frequency of 'https' is a strange phenomenon. After further observation, we found that they might be the URLs of the news in each piece of data. Therefore, in the data preprocessing step, we can consider removing the URLs from the sentences.

\section{Methodology}
We propose two fake news detection models: one is the Text-RNN model based on bidirectional LSTM, and the other is Text-Transformers based on transformers. The description of the two models is as follows.

\subsection{Text-RNN}
\begin{figure}[h]
\centering
\caption{Text-RNN model based on bidirectional LSTM.} \label{figure3}
 \includegraphics[width=8cm,height=10cm ]{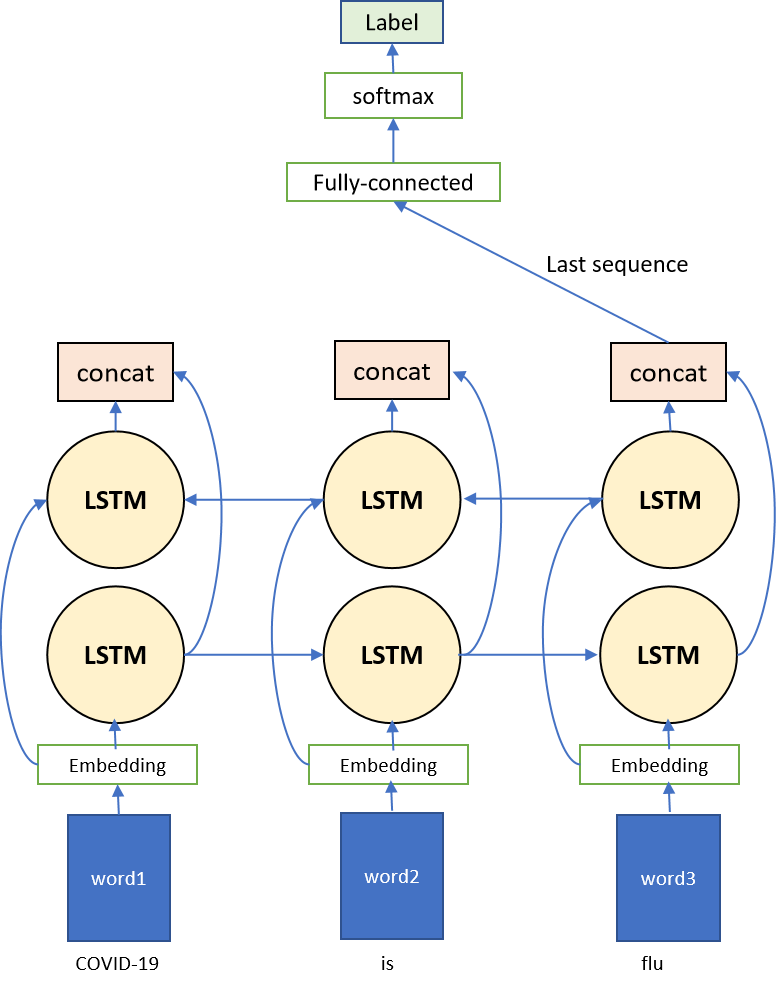}
\end{figure}

Although the LSTM-based deep neural network has proven its effectiveness, but one disadvantage is that the LSTM is based on the previous text information. Therefore, our first model uses a bidirectional LSTM to overcome this shortcoming. The architecture of the model is shown in the Fig. \ref{figure3}.

In the TextRNN model, we use the GloVe~\cite{pennington2014glove} word vector as our embedding layer  with the dimension of 200. After the encoded word vector passes through the bidirectional LSTM, we take the hidden state of the last layer and get the final result through the fully connected layer.

\subsection{Text-transformers}


Contextualized language models such as ELMo and Bert trained on large corpus have demonstrated
remarkable performance gains across many NLP tasks recently. In our
experiments, we use various architectures of language models as the backbone of our second model.

\begin{figure}[h]
\caption{Five-fold Five-model cross-validation framework based on pre-trained language models.} \label{fig1}
\includegraphics[width=\textwidth]{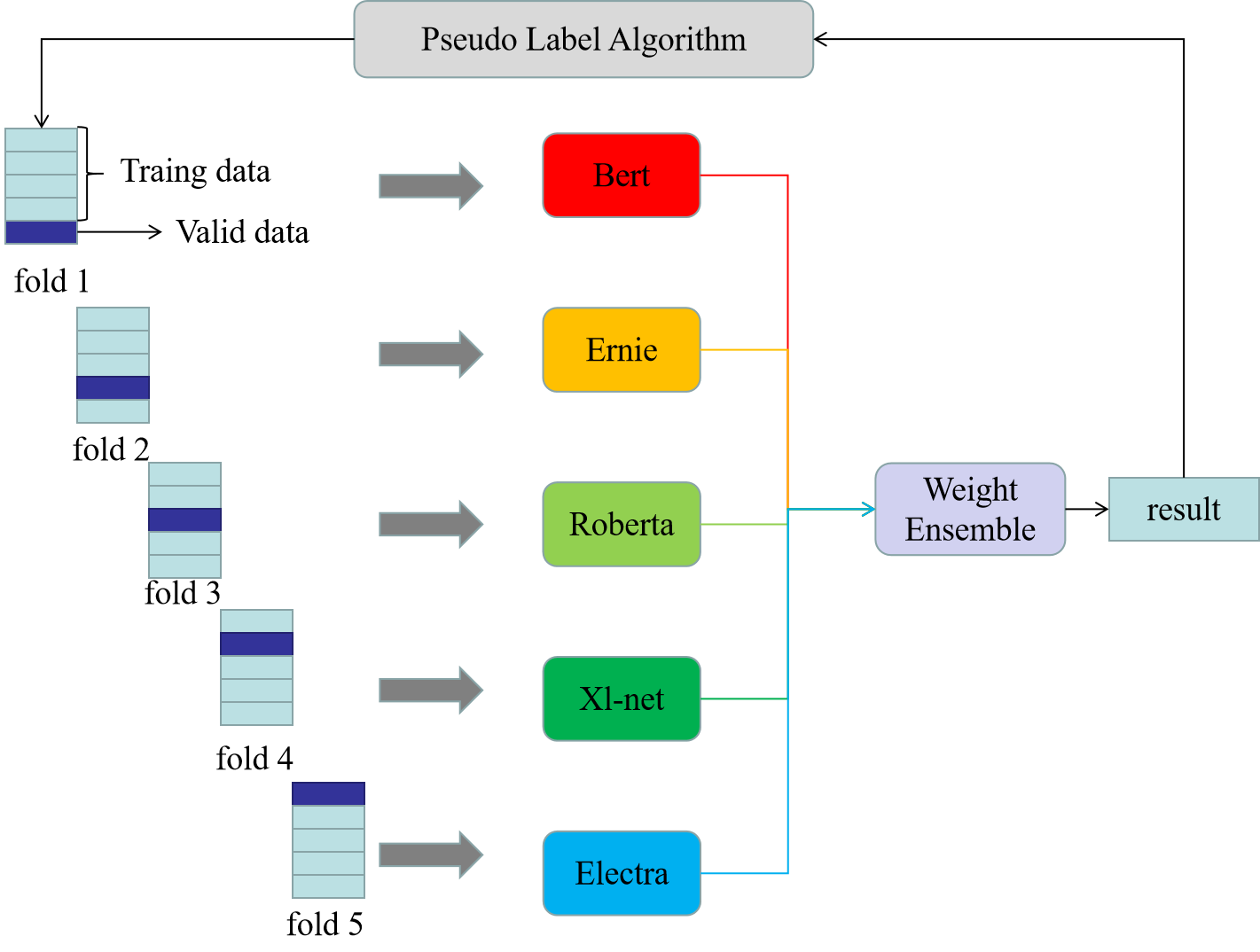}
\end{figure}

As shown in the Fig.4, for the architecture of the language model, we use five different language models including Bert, Ernie, Roberta, XL-net, and Electra trained with the five-fold cross-validation.
We have designed three training methods for this model architecture:
\begin{itemize}
\item \textbf{Five-fold Single-model Ensemble}: For each fold of the five-fold cross-validation method, we use same models for fine-tuning.

\item \textbf{Five-fold Five-model Ensemble}: For each fold of the five-fold cross-validation method, we use different models for fine-tuning.

\item \textbf{Pseudo Label Algorithm}: Because the amount of data is too small, we propose a pseudo-label algorithm to do data augmentation. If a test data is predicted with a probability greater than 0.95, we think that the data is predicted correctly with a relatively high confidence and add it into the training set.

\item \textbf{Weight Ensemble}: We adopt soft voting as an integration strategy, which refers to taking the average of the probabilities of all the models predicted to a certain class as the standard and the type of corresponding with the highest probability as the final prediction result. In our method, we take the highest f1-score of each fold model on the validation set as the ensemble weight.

\end{itemize}

\section{Experiments}
\subsection{Experimental Settings}
\begin{itemize}

\item \textbf{Text-RNN}: The epoch is set to 120, learning rate to 0.01, batch size to 128, text length to 140, and drop out rate to 0.2.
The learning rate is multiplied by the attenuation coefficient 0.1 every 30 epochs.

\item \textbf{Text-Transformers}: The epoch of each fold is set to 12, the batch size is set to 256, the maximum length of the text is set to 140. For the Text-transformers model, due to the complexity of transformer model, we adopt the training strategy as shown in 4.2.
\end{itemize}

\subsection{Training Strategy}
\begin{itemize}

\item \textbf{Label Smoothing}: Label smoothing  \cite{szegedy2016inception} is a regularization technique that introduces noise for the labels. Assuming for a small constant $\epsilon$, the training set label $y$ is correct with a probability or incorrect otherwise. Label Smoothing regularizes a model based on a softmax with  output values by replacing the hard 0 and 1 classification targets with targets of $\frac{\epsilon }{k-1}$ and $1-\epsilon$ respectively.
In our strategy, we take $\epsilon$ equal to 0.01.

\item \textbf{Learning Rate Warm Up}: Using too large learning rate may result in numerical instability especially at the very beginning of the training, where parameters are randomly initialized. The warm up \cite{he2016deep} strategy increases the learning rate from 0 to the initial learning rate linearly during the initial $N$ epochs or $m$ batches. In our strategy, we set an initial learning rate of 1e-6, which increased gradually to 5e-5 after 6 epochs.

\item \textbf{Learning Rate Cosine Decay}: After the learning rate warmup stage described earlier, we typically steadily decrease its value from the initial learning rate.  Compared to some widely used strategies including exponential decay and step decay, the cosine decay \cite{loshchilov2016sgdr} decreases the learning rate slowly at the beginning, and then
becomes almost linear decreasing in the middle, and slows down again at the end. It potentially improves the training progress. In our strategy, after reaching a maximum value of 5e-5, the learning rate decreases to 1e-6 after a cosine decay of 6 epochs

\item \textbf{Domain Pretraining}: Sun  et.  al.~\cite{sun2019fine} demonstrated that pre-trained models such as Bert, which do further domain pretraining on the dataset, can lead to performance gains. Therefore, we adopt \textbf{Covid-Twitter-Bert} which is pretrained on a large corpus of twitter messages on the topic of COVID-19.

\end{itemize}

\subsection{Results}
In Table 2, we presented our results. We evaluated
our models using the official competition metric weighted F1-score which is F1-score averaged across the classes.
\begin{table}[htbp]
\caption{Results of different models.}\label{table2}
\centering
\begin{tabular}{@{}c|c|c|c|c@{}}
\toprule
Method                                                                                                                        & accuracy & precision & recall & weighted F1-score \\ \midrule
TextRNN                                                                                                                       & 0.924    & 0.935     & 0.924   & 0.926             \\ \midrule
\begin{tabular}[c]{@{}c@{}}Text-Transformers\\ +Five-fold single model cross-validation\end{tabular}                          & 0.976    & 0.974     & 0.974   & 0.976            \\ \midrule
\begin{tabular}[c]{@{}c@{}}Text-Transformers\\ +Five-fold five model cross-validation\end{tabular}                            & 0.980    & 0.982     & 0.980   & 0.981             \\ \midrule
\begin{tabular}[c]{@{}c@{}}Text-Transformers\\ + Five-fold five model cross-validation\\ +Pseudo Label Algorithm\end{tabular} & 0.985    & 0.986     & 0.985   & 0.985             \\ \bottomrule
\end{tabular}
\end{table}

\begin{figure}[htbp]
\caption{Results of five-fold five-model ensemble. The blue and orange lines represent val F1 score, train F1 score, and the red and green lines represent val loss and train loss. }\label{figure6}
\centering

\subfigure[Fold 1 for Bert.]{
\begin{minipage}[t]{0.33\linewidth}
\centering
\includegraphics[width=3.5cm,height=2.7cm]{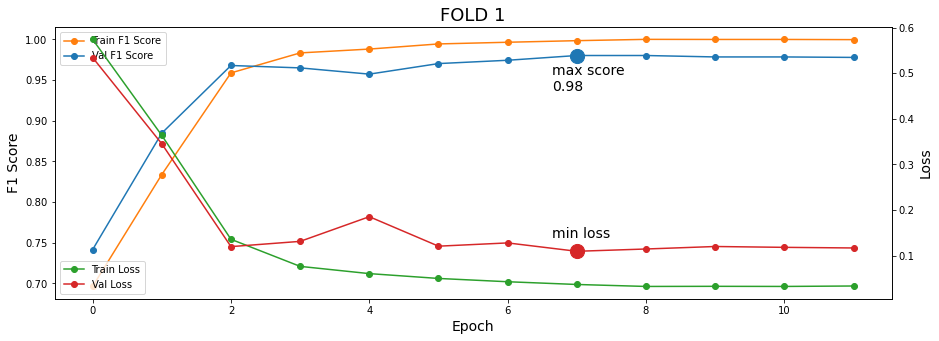}
\end{minipage}%
}%
\subfigure[Fold 2 for Ernie]{
\begin{minipage}[t]{0.33\linewidth}
\centering
\includegraphics[width=3.5cm,height=2.7cm]{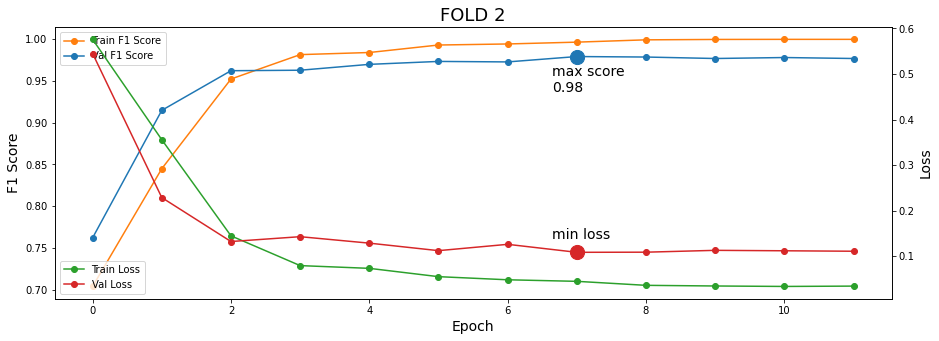}
\end{minipage}%
}%
\subfigure[Fold 3 for Roberta]{
\begin{minipage}[t]{0.33\linewidth}
\centering
\includegraphics[width=3.5cm,height=2.7cm]{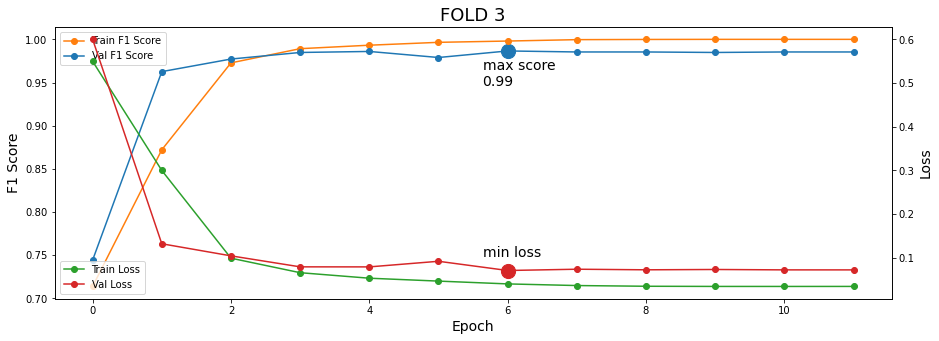}
\end{minipage}
}%

\subfigure[Fold 4 for Xl-net]{
\begin{minipage}[t]{0.33\linewidth}
\centering
\includegraphics[width=3.5cm,height=2.7cm]{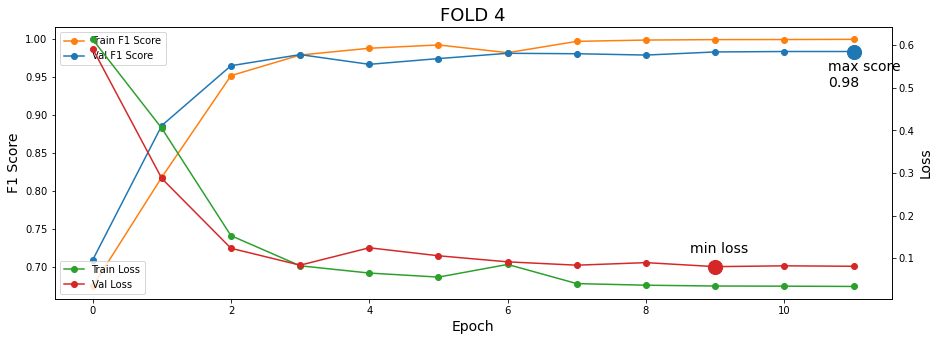}
\end{minipage}
}%
\subfigure[Fold 5 for Electra]{
\begin{minipage}[t]{0.33\linewidth}
\centering
\includegraphics[width=3.5cm,height=2.7cm]{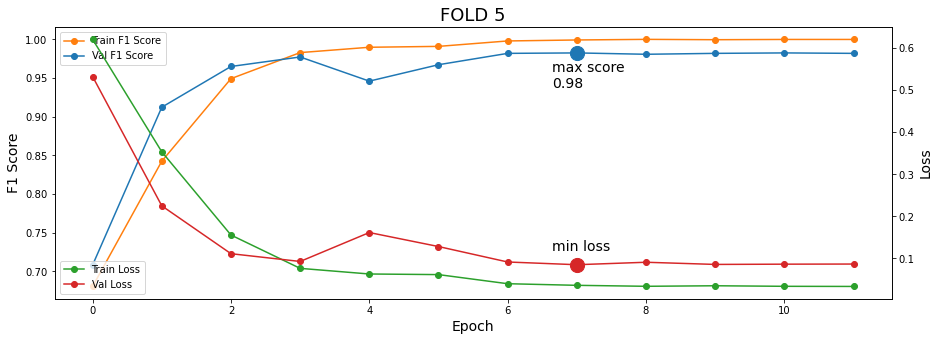}
\end{minipage}
}%

\centering
\end{figure}

\begin{figure}[H]
\centering
\caption{Confusion matrix of predicted result and true label}\label{figure5}
\includegraphics[width=8cm,height = 5cm]{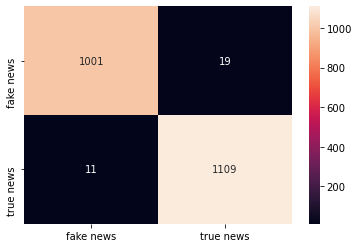}
\end{figure}

In order to make full use of the data, we merged the train set and the valid set. For TextRNN, we re-divided the merged data into the training set and the validation set at a ratio of 8:2, and performed single-fold cross-validation. The weighted f1-score is 0.926.
For the Text-transformers model, we used five-fold cross-validation. Then we compared five-fold single-model cross-validation with five-fold five-model cross-validation. Finally, we achieved the weighted F1 scores of 0.975 and 0.981, respectively. After adding the pseudo-label, the weighted F1 score of 0.985 was obtained on the test set, achieving the third place in the competition which attracted 421 teams to participate in total. Fig.\ref{figure6} shows the performance of our model in each fold.

\subsection{Analysis}

In order to further understand the results on the test set, we investigated the predictions made by our models by conducting simple visualizations of the confusion matrices of predictions acquired by our best models.

From Fig.~\ref{figure5}, we can see that our model has high precision, which is
also obvious from Table \ref{table2} presented above. Fig.~\ref{figure5} also shows that our model has slightly higher false negatives compared to false positives. In other words, the chance of our model mislabeling fake news as true news is slightly higher than predicting true news as fake.

\section{Related Work}

\subsection{Pre-trained Language Models}
Pre-training and then fine-tuning has become a new paradigm in natural language processing. Through self-supervised learning from a large corpus, the language model can learn general knowledge, and then transfer it to downstream tasks by fine-tuning on specific tasks.

Elmo uses Bidirectional LSTM~\cite{hochreiter1997long} to extract word vectors using context information~\cite{peters2018deep}.
GPT~\cite{radford2018improving} enhances context-sensitive embedding by adjusting the transformer~\cite{vaswani2017attention}.
The bidirectional language model BERT~\cite{devlin2018bert} applies cloze and next sentence prediction to self-supervised learning to strengthen word embeddings.  Liu et. al.~\cite{liu2019roberta} removes the next sentence prediction from self-training, and performs  more fully training, getting a better language model na Roberta. Sun et. al.~\cite{sun2019ernie} strengthened the pre-trained language model, completely masking the span in Ernie. Further, Sun et. al.~\cite{sun2020ernie} proposed continuous multi-task pre-training and several pre-training tasks in Ernie 2.0.
 
  In our system, we fine-tuned the above models using the k-fold cross-validation method, which achieved excellent performance.
 
 \subsection{K-fold Cross-Validation}
 $K$-fold cross-validation~\cite{mosteller1968data} means that the training set is divided into $K$ sub samples, one single sub sample is reserved as the data for validation, and the other $K$-1 samples are used for training. Cross-validation is repeated $K$ times, and each sub sample is verified once. The average of the  results or other combination methods are used to obtain a single estimation. The advantage of this method is that it can repeatedly use the randomly generated sub samples for training and verification, and each time the results are verified, the less biased results can be obtained.

The traditional $K$-fold cross-validation uses the same model to train each fold and only retains the best results. In our system, we use different models for each fold and keep the models for each fold to fuse the results. Our experiments prove that this method outperforms the common $K$-fold cross-validation method.

 \subsection{Fake News Detection and Categorization}
 
 In the past few years, there have been several studies of applying computational methods to deal with fake news detection.
Ceron et. al.~\cite{ceron2020fake} used topic models to distinguish fake news, and Hamid et. al.~\cite{hamid2020fake} proposed to use Bag of Words (BoW) and BERT embedding. Yuan et. al.~\cite{yuan2020early} explicitly exploited the credibility of publishers and users for early fake news detection.

However, during the COVID-19 pandemic, it is necessary to establish a reliable automated detection program for COVID-19, but the above-mentioned work rarely studies fake news detection on how to detect COVID-19, and ignores the ensemble strategies of pre-trained language models.

\section{Conclusion}
In this paper, we presented our approach on COVID-19 fake news detection in English. 
We have established two types of models based on bidirectional LSTM and transformer, and the transformer-based model achieved better results in this competition. We proved that five-fold five-model cross-validation performs better than five-fold single-model cross-validation, and pseudo label algorithm can effectively improve the performance. In the future, we plan to use generative models such as T5~\cite{raffel2019exploring} to generate labels directly, further enhancing the predicted results.

\section*{Acknowledgements}
This work was partially supported by National Key Research and Development Project
(2019YFB1704002) and National Natural Science Foundation of China (61876009).

%
%

\bibliographystyle{splncs04} %
\bibliography{reference} %

\begin{thebibliography}{10}
\providecommand{\url}[1]{\texttt{#1}}
\providecommand{\urlprefix}{URL }
\providecommand{\doi}[1]{https://doi.org/#1}

\bibitem{ceron2020fake}
Ceron, W., de~Lima-Santos, M.F., Quiles, M.G.: Fake news agenda in the era of
  covid-19: Identifying trends through fact-checking content. Online Social
  Networks and Media p. 100116 (2020)

\bibitem{devlin2018bert}
Devlin, J., Chang, M.W., Lee, K., Toutanova, K.: Bert: Pre-training of deep
  bidirectional transformers for language understanding. arXiv preprint
  arXiv:1810.04805  (2018)

\bibitem{hamid2020fake}
Hamid, A., Shiekh, N., Said, N., Ahmad, K., Gul, A., Hassan, L., Al-Fuqaha, A.:
  Fake news detection in social media using graph neural networks and nlp
  techniques: A covid-19 use-case (2020)

\bibitem{he2016deep}
He, K., Zhang, X., Ren, S., Sun, J.: Deep residual learning for image
  recognition. In: Proceedings of the IEEE conference on computer vision and
  pattern recognition. pp. 770--778 (2016)

\bibitem{hochreiter1997long}
Hochreiter, S., Schmidhuber, J.: Long short-term memory. Neural computation
  \textbf{9}(8),  1735--1780 (1997)

\bibitem{liu2019roberta}
Liu, Y., Ott, M., Goyal, N., Du, J., Joshi, M., Chen, D., Levy, O., Lewis, M.,
  Zettlemoyer, L., Stoyanov, V.: Roberta: A robustly optimized bert pretraining
  approach. arXiv preprint arXiv:1907.11692  (2019)

\bibitem{loshchilov2016sgdr}
Loshchilov, I., Hutter, F.: Sgdr: Stochastic gradient descent with warm
  restarts. arXiv preprint arXiv:1608.03983  (2016)

\bibitem{mosteller1968data}
Mosteller, F., Tukey, J.W.: Data analysis, including statistics. Handbook of
  social psychology  \textbf{2},  80--203 (1968)

\bibitem{patwa2020fighting}
Patwa, P., Sharma, S., PYKL, S., Guptha, V., Kumari, G., Akhtar, M.S., Ekbal,
  A., Das, A., Chakraborty, T.: Fighting an infodemic: Covid-19 fake news
  dataset. arXiv preprint arXiv:2011.03327  (2020)

\bibitem{pennington2014glove}
Pennington, J., Socher, R., Manning, C.D.: Glove: Global vectors for word
  representation. In: Proceedings of the 2014 conference on empirical methods
  in natural language processing (EMNLP). pp. 1532--1543 (2014)

\bibitem{peters2018deep}
Peters, M.E., Neumann, M., Iyyer, M., Gardner, M., Clark, C., Lee, K.,
  Zettlemoyer, L.: Deep contextualized word representations. arXiv preprint
  arXiv:1802.05365  (2018)

\bibitem{radford2018improving}
Radford, A., Narasimhan, K., Salimans, T., Sutskever, I.: Improving language
  understanding by generative pre-training (2018)

\bibitem{raffel2019exploring}
Raffel, C., Shazeer, N., Roberts, A., Lee, K., Narang, S., Matena, M., Zhou,
  Y., Li, W., Liu, P.J.: Exploring the limits of transfer learning with a
  unified text-to-text transformer. arXiv preprint arXiv:1910.10683  (2019)

\bibitem{sun2019fine}
Sun, C., Qiu, X., Xu, Y., Huang, X.: How to fine-tune bert for text
  classification? In: China National Conference on Chinese Computational
  Linguistics. pp. 194--206. Springer (2019)

\bibitem{sun2020ernie}
Sun, Y., Wang, S., Li, Y.K., Feng, S., Tian, H., Wu, H., Wang, H.: Ernie 2.0: A
  continual pre-training framework for language understanding. In: AAAI. pp.
  8968--8975 (2020)

\bibitem{sun2019ernie}
Sun, Y., Wang, S., Li, Y., Feng, S., Chen, X., Zhang, H., Tian, X., Zhu, D.,
  Tian, H., Wu, H.: Ernie: Enhanced representation through knowledge
  integration. arXiv preprint arXiv:1904.09223  (2019)

\bibitem{szegedy2016inception}
Szegedy, C., Ioffe, S., Vanhoucke, V., Alemi, A.: Inception-v4,
  inception-resnet and the impact of residual connections on learning. arXiv
  preprint arXiv:1602.07261  (2016)

\bibitem{vaswani2017attention}
Vaswani, A., Shazeer, N., Parmar, N., Uszkoreit, J., Jones, L., Gomez, A.N.,
  Kaiser, {\L}., Polosukhin, I.: Attention is all you need. In: Advances in
  neural information processing systems. pp. 5998--6008 (2017)

\bibitem{yuan2020early}
Yuan, C., Ma, Q., Zhou, W., Han, J., Hu, S.: Early detection of fake news by
  utilizing the credibility of news, publishers, and users based on weakly
  supervised learning (2020)

\end{thebibliography}





\end{document}